\documentclass[11pt]{article}

\usepackage{booktabs}
\usepackage{amsmath}
\usepackage{multirow}
\usepackage{algorithm}
\usepackage{algpseudocode}
\usepackage[most]{tcolorbox}
\usepackage{listings}
\usepackage{subcaption}

\newtcblisting{promptbox}{
    listing only,
    breakable,
    colback=gray!5,
    colframe=gray!60,
    boxrule=0.4pt,
    arc=2mm,
    left=2mm,
    right=2mm,
    top=2mm,
    bottom=2mm,
    listing options={
        basicstyle=\ttfamily\footnotesize,
        breaklines=true,
        columns=fullflexible,
        keepspaces=true,
        showstringspaces=false
    }
}

\usepackage[preprint]{acl}

\usepackage{times}
\usepackage{latexsym}

\usepackage[T1]{fontenc}

\usepackage[utf8]{inputenc}

\usepackage{microtype}

\usepackage{inconsolata}

\usepackage{graphicx}

\newcommand{\ignore}[1]{}
 \newcounter{todocounter}

\title{ReLTEx: Reliable LLM-based Taxonomy Expansion}

\author{
  \textbf{Zeinab Ghamlouch\textsuperscript{1}} \quad
  \textbf{Mehwish Alam\textsuperscript{1}}
  \\
  \\
  \textsuperscript{1}Télécom Paris, Institut Polytechnique de Paris, France
  \\
  \small{
    \texttt{zeinab.b.ghamlouch@gmail.com} \quad
    \texttt{mehwish.alam@telecom-paris.fr}
  }
}

\begin{document}
\maketitle
\begin{abstract}
Recent advances in Large Language Models (LLMs) have demonstrated strong capabilities in generating semantically relevant concepts and relations, making them promising tools for taxonomy enrichment. However, directly relying on LLM-generated expansions often leads to noisy, redundant, or hierarchically inconsistent structures, limiting their reliability for automated taxonomy expansion. In this paper, we present \textit{ReLTEx}, a framework for reliable LLM-based taxonomy expansion. ReLTEx combines LLM-driven candidate generation with structure-aware validation and recursive expansion control to improve the consistency and quality of generated taxonomies by reducing hallucinations. We evaluate the proposed framework using benchmark taxonomies under a masked taxonomy expansion setting and compare multiple validation strategies. Experimental results, supported by both adapted evaluation metrics and human evaluation, demonstrate that ReLTEx produces more reliable and semantically coherent taxonomy expansions. 
\end{abstract}

\section{Introduction}
\label{sec:introduction}

Taxonomies provide structured hierarchical representations of knowledge and play a fundamental role in numerous applications, including knowledge graphs~\cite{suchanek2024yago}, semantic search and question answering~\cite{navigli2010babelnet}, etc. By organizing concepts as parent--child relations, taxonomies enable semantic understanding, efficient navigation, and knowledge discovery. However, as domains evolve and new concepts continuously emerge, maintaining and expanding taxonomies manually becomes increasingly costly, time-consuming, and difficult to scale~\cite{navigli2010babelnet}.

To address these challenges, extensive research has explored automated taxonomy expansion. Early approaches relied on lexico-syntactic patterns~\cite{hearst1992automatic}, followed by distributional semantic methods~\cite{zhang2018taxogen} and graph-based techniques~\cite{pietrasik2024non}. Moreover, Pretrained Language Models (PLMs) have substantially improved taxonomy expansion by learning contextual representations of concepts and hierarchical relations~\cite{liu2021temp}. Despite these advances, automated taxonomy expansion remains challenging. A generated concept may be \emph{semantically ambiguous}, as in \textit{``Bank''}, which can refer to either a financial institution or the side of a river. Candidate generation may also introduce \emph{noise or redundancy}, for example by proposing both \textit{``Question''} and \textit{``Inquiry''}. Moreover, an otherwise valid concept may be \emph{attached to an inappropriate parent}, such as placing \textit{``Answer''} under \textit{``CreativeWork''} rather than \textit{``Comment''}. These errors can compromise hierarchical coherence and propagate during recursive expansion.

Recent advances in Large Language Models (LLMs) have created new opportunities for taxonomy enrichment which uses prompting for taxonomy construction~\cite{zeng2024chain} and hierarchical attachment prediction~\cite{mishra2024flame}. Because of their extensive parametric knowledge and strong generative capabilities, LLMs can infer hierarchical relations and propose semantically relevant concepts in a zero-shot setting, making them attractive for expanding existing taxonomies, particularly in domains where curated resources are scarce or rapidly evolving. However, without explicit structural verification, LLM-generated concepts may violate the taxonomy hierarchy, and such errors can accumulate during recursive expansion.

To address these challenges, we propose \textit{ReLTEx\footnote{The source code and the benchmark splits are available at \url{https://github.com/zeinabGhamlouch/ReLTEx}.}}, a framework for reliable LLM-based taxonomy expansion. ReLTEx combines zero-shot LLM-based candidate generation with structure-aware validation and recursive expansion control. Candidate parent--child relations are verified using a path-aware classifier trained on taxonomy structures, allowing structurally inconsistent generations to be filtered before insertion into the taxonomy. Recursive stopping criteria further regulate recursive expansion and reduce the propagation of erroneous generations.

We evaluate ReLTEx on benchmark taxonomies under a masked taxonomy expansion setting, where hidden concepts must be recovered from partial taxonomies. Our evaluation consists of standard and adapted evaluation metrics, human assessment, and LLM based assessment along with an ablation study. The main contributions of this work are as follows, we propose:

\begin{itemize}
\item A zero-shot LLM-based taxonomy expansion framework that exploits contextual information from the taxonomy hierarchy to generate candidate concepts.

\item A structure-aware validation mechanism that mitigates hallucinated and structurally inconsistent parent--child relations, improving the reliability of recursive taxonomy expansion.

\item A comprehensive evaluation protocol based on adapted evaluation metrics, human evaluation, and LLM based assessment.
\end{itemize}

This paper is organized as follows. Section~\ref{sec:related_work} reviews related work while Section~\ref{sec:problem_formulation} defines the taxonomy expansion problem. Section~\ref{sec:methodology} presents ReLTEx. Section~\ref{sec:experimental-setup} describes the datasets and evaluation metrics, and Section~\ref{sec:results} presents the experimental results. Finally, Section~\ref{sec:conclusion} concludes the paper.

\section{Related Work}
\label{sec:related_work}

Recent methods formulate taxonomy expansion as \emph{node attachment problem}, where predefined candidate concepts are inserted into an existing taxonomy by predicting their most appropriate parent node. In the following we categorize these methods into structure based and LLM-based methods.

\paragraph{Structure based Methods.} TaxoExpan \cite{shen2020taxoexpan} leverages Graph Neural Networks and self-supervised learning to insert unseen concepts through ego-graph representations. Building on the use of structural information, TEMP \cite{liu2021temp} models taxonomy paths using PLMs and optimizes a dynamic margin ranking objective over positive and negative insertion paths. % rather than relying on a single representation. 
STEAM \cite{yu2020steam} adopts a multi-view co-training framework that combines distributed, contextual, and lexico-syntactic representations through mini-path structures. Other approaches have similarly focused on improving hierarchical attachment through graph-based representations~\cite{shang2020taxonomy}, contrastive learning strategies~\cite{niu2024contrastive}, or PLM embeddings~\cite{takeoka2021low}. 

Despite their effectiveness, most existing taxonomy expansion approaches \cite{shen2020taxoexpan, margiotta2023taxosbert} operate under a constrained setting in which the candidate concepts are provided in advance instead of generating entirely new concepts dynamically.

\paragraph{LLM-based Methods.} %Recent work has explored several uses of LLMs in taxonomy-related tasks.
TaxoGlimpse~\cite{sun2024large} assesses the taxonomic knowledge encoded in LLMs across general and specialized taxonomies. The results show that the performance of an LLM deteriorates for specialized domains and deeper taxonomy levels. However, their work focuses on evaluating taxonomic knowledge rather than expanding an existing hierarchy. %Other methods use language models for taxonomy construction or attachment prediction. 
Chain-of-Layer~\cite{zeng2024chain} iteratively induces a taxonomy from a given set of entities through layer-wise prompting and an ensemble-based relation filtering mechanism. FLAME~\cite{mishra2024flame} leverages LLMs to predict the most appropriate parent for a query concept that is known in advance. 

Taxoria~\cite{ghamlouch2025enriching} is the closest work to ours, as it recursively generates novel concepts based on an existing taxonomy without relying on a predefined candidate pool. However, its validation is based primarily on embedding-based semantic similarity and does not explicitly model the structural compatibility of a generated parent--child relation. % using its hierarchical path. 
Consequently, erroneous or weakly related generations may be accepted and propagated during recursive expansion.

Unlike prior taxonomy expansion methods that either attach predefined concepts or rely primarily on semantic similarity, ReLTEx combines open-ended LLM generation with path-aware structural validation and recursive expansion control. By validating each generated parent--child relation in its hierarchical context before insertion.

\section{Problem Formulation} \label{sec:problem_formulation}

A taxonomy can be represented as \(T=(V,E)\), where \(V\) denotes the set of concepts (nodes) and \(E\) represents parent--child relations between the concepts. Given an initial seed taxonomy \(T_0=(V_0,E_0)\), the goal of taxonomy expansion is to enrich the taxonomy with semantically relevant concepts while preserving hierarchical consistency. Formally, given a node \(v \in V_0\) and its local hierarchical context, the objective is to generate a set of candidate child concepts $C_v=\{c_1,c_2,\ldots,c_k\}$. Where \(k\) denotes the number of candidate child concepts generated for node \(v\), and \(c_i\) is a potential child concept of \(v\), where \(i=1,\ldots,k\).

The generated candidates may be hallucinated, redundant, or structurally inconsistent. Thus, the goal is to select a curated subset suitable for safe taxonomy integration. The resulting expanded concept set can therefore be expressed as $V' = V_0 \cup A_v$; where \(A_v \subseteq C_v\) denotes the subset of generated candidates selected for insertion into the taxonomy. 
\section{ReLTEx}
\label{sec:methodology}

\begin{figure*}
\centering
\includegraphics[width=0.9\linewidth]{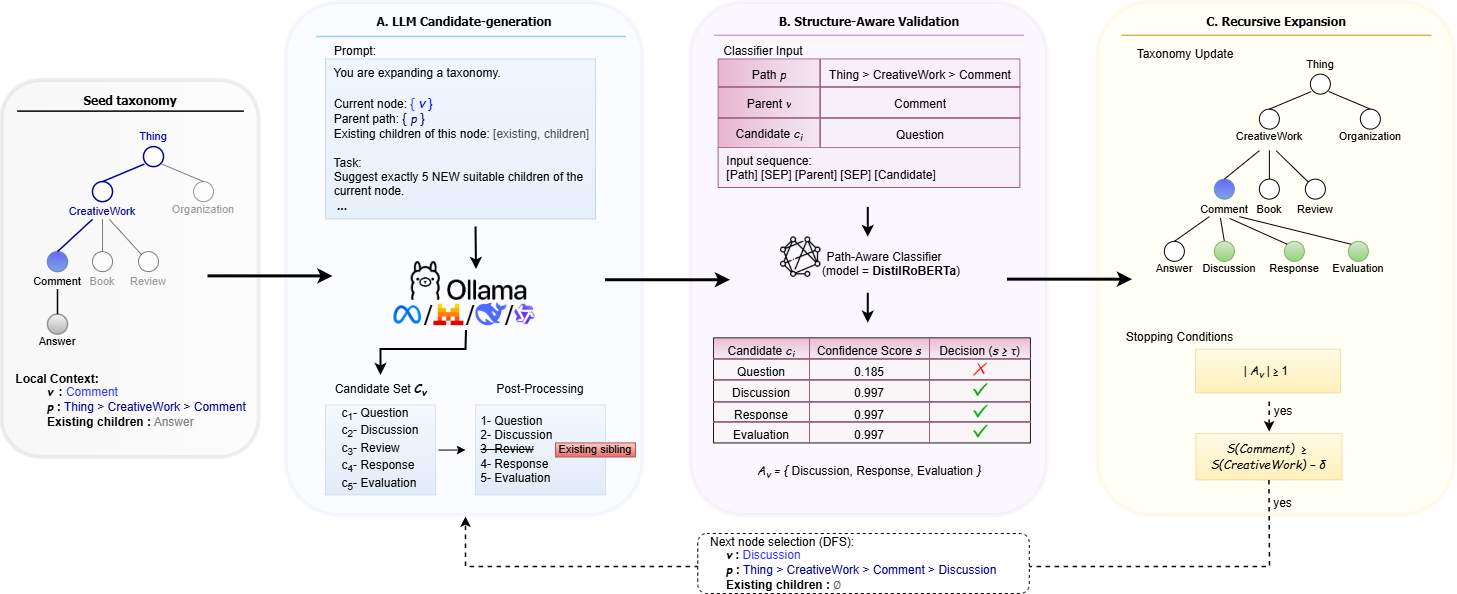}
\caption{Overview of the ReLTEx framework.}
\label{fig:pipeline}
\vspace{-5mm}
\end{figure*}

Figure~\ref{fig:pipeline} shows an overview of ReLTEx framework. Starting from a seed taxonomy, ReLTEx expands each taxonomy node in three stages: \textbf{(A)} LLM-based candidate generation, \textbf{(B)} structure-aware validation, and \textbf{(C)} recursive expansion using a stopping mechanism. The following subsections describe each stage in detail.

\subsection{LLM-based Candidate Generation}

ReLTEx performs recursive taxonomy expansion through a depth-first traversal of the taxonomy. For each target node, the framework constructs a prompt using its local hierarchical context, consisting of the path from the taxonomy root to the target node together with its existing children. The ancestor path provides the semantic context of the current taxonomy branch, while the existing children illustrate the desired level of abstraction, guiding the LLM to generate semantically relevant child concepts at the appropriate level of granularity. The language model is instructed to generate new child concepts considering the appropriate granularity level while avoiding duplicates, synonyms, and simple rephrasing of the existing or newly generated siblings (Appendix~\ref{sec:appendix} shows the prompt). 
For each parent node, the LLM is prompted to generate a fixed number \(k\) of candidate child concepts. The parameter \(k\) controls the branching factor of the expanded taxonomy and can be adjusted depending on the desired level of expansion. After generation, candidate names undergo basic lexical normalization. Exact lexical duplicates under this canonical representation, as well as candidates matching existing children or sibling nodes, are removed before the remaining candidates are passed to the validation stage.

\subsection{Structure-Aware Candidate Validation}

We investigated three validation strategies: (i) semantic similarity filtering, which measures the semantic compatibility between the generated concept and its local taxonomy context; (ii) LLM-based validation, where an LLM judges whether the generated concept represents a valid, non-redundant child of the parent; and (iii) classifier-guided validation, which predicts the structural validity of the parent--child relation using a classifier trained on taxonomy edges. In this subsection we focus on classifier guided validation.

Unlike semantic similarity measures, the proposed classifier learns structural compatibility from annotated taxonomy relations by jointly considering the hierarchical context, the parent concept, and the generated child concept. 
The validation module is formulated as a binary classification task using a DistilRoBERTa encoder fine-tuned on labeled parent--child relation pairs. The classifier is trained on positive and negative examples derived from the seed taxonomy, enabling it to discriminate between valid and invalid hierarchical relations.
Given the hierarchical path \(p\), the parent concept \(v\), and a generated candidate child concept \(c_i\), represented as \texttt{Path [SEP] Parent [SEP] Child}, the classifier outputs a confidence score
\[
s(p,v,c_i)\in[0,1],
\]
defined as the softmax probability assigned to the positive class, representing the likelihood that the relation \((v,c_i)\) corresponds to a valid taxonomy edge within the hierarchical context \(p\). Candidate concepts with \(s(p,v,c_i)\geq\tau\), where \(\tau\) denotes the validation threshold, are accepted and incorporated into the taxonomy; otherwise, they are discarded.

Positive examples correspond to valid taxonomy edges, whereas negative examples are generated automatically using hierarchy-aware perturbations that produce semantically plausible but structurally incorrect relations. %Using the seed taxonomy illustrated in Figure~\ref{fig:pipeline} as a running example, 
We construct the following hard negative examples:

\begin{itemize}
    \item \textbf{Reversed edges}, obtained by reversing a valid parent--child relation, e.g., \textit{Comment} $\rightarrow$ \textit{Answer} becomes \textit{Answer} $\rightarrow$ \textit{Comment} (see Figure~\ref{fig:pipeline} for running example).
    
    \item \textbf{Sibling confusions}, where the correct parent is replaced by one of its siblings, e.g., \textit{Comment} $\rightarrow$ \textit{Answer} becomes \textit{Book} $\rightarrow$ \textit{Answer}.
    
    \item \textbf{Grandparent--child confusions}, where the immediate parent is replaced by its parent, e.g., \textit{Comment} $\rightarrow$ \textit{Answer} becomes \textit{CreativeWork} $\rightarrow$ \textit{Answer}.
    
    \item \textbf{Same-depth mismatches}, where the child is attached to a concept at the same hierarchical level as its true parent, e.g., \textit{Article} $\rightarrow$ \textit{Answer}.
    
    \item \textbf{Nearby hierarchy confusions}, where the correct parent is replaced by a semantically related concept from a neighboring branch, e.g., \textit{Review} $\rightarrow$ \textit{Answer}.
    
    \item \textbf{Random invalid relations}, obtained by randomly pairing unrelated concepts, e.g., \textit{Organization} $\rightarrow$ \textit{Answer}.
\end{itemize}

Each training instance is represented as \texttt{Path [SEP] Parent [SEP] Child}, where the path corresponds to the ancestor sequence from the taxonomy root to the parent concept. By training on both positive examples and hierarchy-aware hard negatives, the classifier learns structural compatibility rather than relying solely on semantic similarity.

\subsection{Recursive Expansion}

Recursive generative expansion may progressively introduce semantic drift and uncontrolled taxonomy growth. For example, once an incorrect parent--child relation is accepted, subsequent expansion may continue from that erroneous node and generate concepts that increasingly diverge from the intended taxonomy branch. To mitigate this issue, ReLTEx reuses the confidence scores assigned by the classifier during child validation to regulate recursive expansion at the branch level.

Only the concepts accepted during the validation step are eligible for further recursive expansion. Let \(A_v\) denote the set of accepted child concepts generated for node \(v\):
\begin{equation*}
A_v=\{c_i\in C_v \mid s(p,v,c_i)\geq\tau\}
\end{equation*}
ReLTEx first requires a minimum number of accepted children: $|A_v| \geq m,$ where \(m\) is the minimum number of accepted children required to continue expanding the branch. If \(|A_v|<m\), expansion at node \(v\) terminates. If \(m=1\), a branch stops when none of its generated candidates passes the validation threshold.

When \(|A_v|\geq m\), ReLTEx estimates the overall reliability of the current expansion step using the mean confidence of the accepted children:

\begin{equation*}
S(v)=\frac{1}{|A_v|}\sum_{c_i\in A_v}s(p,v,c_i).
\end{equation*}

Averaging considers all accepted children and provides a normalized estimate of the reliability of the expansion step, without being determined by a single exceptionally high- or low-confidence prediction.

If we have \(S(v)\) and \(S(\mathrm{parent}(v))\) (mean confidence of its parent node), the expansion below \(v\) continues only if

\begin{equation*}
S(v) \geq S(\mathrm{parent}(v))-\delta,
\end{equation*}

where \(\delta\) is the maximum allowable decrease in average classifier confidence between two consecutive expansion levels. For the initial expansion step, where no parent confidence score is available, this criterion is not applied. This recursive stopping mechanism recursively expands only branches whose confidence remains sufficiently stable.
\section{Experimental Setup} \label{sec:experimental-setup}

\subsection{Datasets and LLMs}

We evaluate ReLTEx on two taxonomy benchmarks that differ substantially in size and structure: \textbf{(i)} \textit{SemEval-2016 Task 13 Environment} taxonomy~\cite{bordea-etal-2016-semeval}, a standard benchmark for taxonomy expansion; \textbf{(ii)} \textit{Schema.org}\footnote{\url{https://schema.org/}}, a large-scale real-world taxonomy that covers a broad range of domains and semantic concepts.
Dataset statistics are summarized in Table~\ref{tab:dataset-summary}.

\begin{table}[ht]
\centering
\small
\caption{Statistics of the taxonomy benchmarks. $|N|$ denotes \#nodes, $|E|$ denotes \#edges, and $|D|$ the maximum depth (with the root counted as level~1).}
\setlength{\tabcolsep}{4pt}
\begin{tabular}{l c c c}
\toprule
\textbf{Dataset} & $|N|$ & $|E|$ & $|D|$ \\
\midrule
SemEval-env & 263 & 262 & 6 \\
Schema.org & 1,143 & 1,142 & 6 \\
\bottomrule
\end{tabular}
\label{tab:dataset-summary}
\end{table}

Experiments are conducted using four open-source language models from different families: \textit{Llama3.2:3B, Mistral:7B, Qwen3:8B, and DeepSeek-R1:8B}. These models were selected to cover a diverse range of capabilities, including lightweight instruction following, general-purpose generation, enhanced reasoning, and explicit reasoning via a thinking model. 

\subsection{Evaluation Metrics}
Our evaluation builds upon metrics commonly adopted in taxonomy expansion literature~\cite{liu2021temp, zhang2024peb}, including Recall@K, Mean Reciprocal Rank (MRR), and the Wu \& Palmer similarity measure. Unlike traditional taxonomy expansion, ReLTEx performs generative expansion, where language models may produce multiple candidate concepts, semantically equivalent variants, or valid concepts attached under alternative parent nodes. To account for these characteristics, we adapt the classical evaluation metrics to the generative setting while preserving their original objectives. Specifically, we distinguish between concept recovery and structural attachment quality, and additionally incorporate semantic matching to account for lexical variability in generated concepts.

The primary quantitative evaluation follows a masked taxonomy expansion protocol (described in Section~\ref{sec:results}), where a subset of taxonomy nodes is hidden and treated as ground truth. The enriched taxonomy is generated from the remaining seed taxonomy, and the recovered concepts are evaluated using the metrics defined below.

Let \(Y\) denote the set of hidden concepts, with \(n = |Y|\). For each hidden concept \(y \in Y\), let \(v_y\) denote the original parent node from which \(y\) was masked. Let \(rank(y)\) denote the ranking position of \(y\) within the accepted candidate set \(A_{v_y}\) (see Section~\ref{sec:methodology} for details).

We define the set of locally recovered hidden concepts as
\[
\Gamma_{\mathrm{local}}
=
\left\{
y \in Y
\;\middle|\;
y \in A_{v_y}
\right\},
\]
where a hidden concept is considered recovered if it appears among the accepted generated children of its original parent node.

\paragraph{Recall@K (R@K)} measures the proportion of hidden concepts that are successfully recovered under their original parent nodes:
\[
R@K
=
\frac{|\Gamma_{\mathrm{local}}|}{n}.
\]

\paragraph{SoftRecall@K (SR@K).}

Exact lexical matching may underestimate taxonomy expansion quality when semantically equivalent concepts are generated. To account for this, we introduce SR@K.

\[
SR@K(y)
=
\begin{cases}
1,
&
\text{if }
\displaystyle
\max_{c_i \in A_{v_y}}
\operatorname{sim}(y,c_i)
\geq
\rho,
\\
0,
&
\text{otherwise.}
\end{cases}
\]

where $y \in Y$ is the hidden concept, \(\operatorname{sim}(\cdot,\cdot)\) denotes cosine similarity between embeddings generated by the \texttt{BAAI/bge-small-en-v1.5} model~\cite{bge_embedding}. We set \(\rho=0.85\), which provides a conservative balance between recognizing semantically equivalent concepts while avoiding matches between only loosely related concepts. The final SR@K score is computed as

\[
SR@K
=
\frac{1}{n}
\sum_{y\in Y}
SR@K(y).
\]

\paragraph{Mean Reciprocal Rank (MRR).}

Accepted candidate concepts are ranked according to decreasing classifier confidence score \(s(p,v,c_i)\). We report two complementary variants of MRR: (i) \(MRR_{\mathrm{overall}}\) and (ii) \(MRR_{\mathrm{found}}\). \(MRR_{\mathrm{overall}}\) measures ranking quality over all hidden concepts. Recovered concepts contribute the reciprocal of their ranking position, while hidden concepts that are not recovered receive a score of zero, thereby penalizing failure to recover hidden concepts:

\[
MRR_{\mathrm{overall}}
=
\frac{1}{n}
\sum_{y\in Y}
\begin{cases}
\dfrac{1}{rank(y)},
&
\text{if } y\in A_{v_y},
\\
0,
&
\text{otherwise.}
\end{cases}
\]

 \(MRR_{\mathrm{found}}\) evaluates ranking quality only for successfully recovered hidden concepts:

\[
MRR_{\mathrm{found}}
=
\frac{1}{|\Gamma_{\mathrm{local}}|}
\sum_{y\in\Gamma_{\mathrm{local}}}
\frac{1}{rank(y)}.
\]

\paragraph{Wu \& Palmer (WuP) Similarity.}

To evaluate the structural quality of node generation, we adapt the WuP similarity to compare predicted parent placements against the original taxonomy structure. For each hidden concept \(y\in Y\), let

\[
V_y^{\mathrm{pred}}
=
\{
v
\mid
y\in A_v
\}
\]
denote the set of parent nodes under which a hidden concept $y$ was generated. The attachment similarity score is then defined as
\[
w(y)
=
\max_{v \in V_y^{\mathrm{pred}}}
\frac{
2 \cdot depth(LCA(v,v_y))
}{
depth(v)+depth(v_y)
}.
\]
where \(LCA(\cdot,\cdot)\) denotes the Lowest Common Ancestor. Higher similarity values indicate that the predicted parent node is structurally closer to the original parent node. 
We further define the set of globally recovered hidden concepts as

\[
\Gamma_{\mathrm{global}}
=
\{
y\in Y
\mid
V_y^{\mathrm{pred}}\neq\emptyset
\}.
\]

We report two variants of the WuP score. The first variant measures attachment quality over all hidden concepts. Successfully generated concepts contribute their WuP score, while hidden concepts that are not successfully generated are scored zero.
\[
WuP_{\mathrm{overall}}
=
\frac1n
\sum_{y\in Y}
\begin{cases}
w(y),
&
\text{if }
y\in\Gamma_{\mathrm{global}},
\\
0,
&
\text{otherwise.}
\end{cases}
\]

The second variant evaluates attachment quality only for successfully generated concepts:

\[WuP_{\mathrm{found}}
=
\frac1{|\Gamma_{\mathrm{global}}|}
\sum_{y\in\Gamma_{\mathrm{global}}}
w(y).\]
\section{Evaluation Results} \label{sec:results}
We evaluate ReLTEx using three complementary protocols: (i) Masked Taxonomy Expansion, (ii) Human Evaluation, (iii) LLM based taxonomy evaluation. .

\subsection{Masked Taxonomy Expansion Benchmark}
We evaluate concept recovery using a masked taxonomy expansion benchmark, following the standard evaluation protocol~\cite{xu2024fuse}. We remove 20\% of the nodes from the seed taxonomy and treat them as hidden test concepts. The remaining taxonomy serves as the seed taxonomy provided to ReLTEx, while the hidden concepts constitute the gold standard used for evaluation. 
Only leaf nodes are considered eligible for masking in order to preserve the overall taxonomy structure. % during evaluation.

Masking is performed under two constraints. \textbf{(i)} Hidden nodes are selected only from those parent nodes retaining at least one visible child after masking, preventing the removal of all local structural context. \textbf{(ii)} A maximum of five hidden children is allowed under any single parent node. These constraints prevent a small number of high-degree parent nodes from dominating the evaluation process and ensures a more balanced distribution of hidden concepts across the taxonomy.

The generator LLM\footnote{Unless otherwise specified, candidate generation is performed using Mistral:7B.} produces \(k=5\) candidate child concepts for each parent node. The root node is excluded from expansion. Generated parent--child candidates are then validated. 
Candidates are accepted only if their confidence score $\geq$ 0.90.
Although ReLTEx supports recursive taxonomy expansion, benchmark experiments are performed exclusively on the masked seed taxonomy. Newly generated concepts are not recursively expanded. 

Since existing taxonomy expansion methods assume a predefined candidate set rather than generating new concepts, no directly comparable baseline exists for our setting.
Table~\ref{tab:benchmark_results} summarizes the benchmark results under the masked taxonomy expansion setting on the SemEval Environment and Schema.org taxonomies. The higher numbers (shown in bold) indicate better performance or quality in the rest of this paper.  

\begin{table}[t]
\centering
\scriptsize
\caption{Results of the masked taxonomy expansion benchmark.  }
\label{tab:benchmark_results}
\begin{tabular}{lcccc}
\toprule
\textbf{Metric}
& \textbf{Llama3.2}
& \textbf{Mistral}
& \textbf{Qwen3}
& \textbf{DeepSeek}\\
% & \textbf{GPT-oss:20b}\\
\midrule

\multicolumn{5}{c}{\textbf{SemEval Environment}} \\
\midrule

R@K
& \textbf{44.23 $\pm$ 2.11}
& 42.31 $\pm$ 1.22
& 42.95 $\pm$ 0.99
& 28.85 $\pm$ 6.20 \\

SR@K
& \textbf{53.21 $\pm$ 2.63}
& 46.47 $\pm$ 1.89
& 47.76 $\pm$ 2.25
& 40.70 $\pm$ 4.78 \\

MRR$_o$
& 23.47 $\pm$ 3.10
& \textbf{23.81 $\pm$ 0.52}
& 21.69 $\pm$ 1.08
& 14.53 $\pm$ 2.61 \\

MRR$_f$
& 52.92 $\pm$ 5.02
& \textbf{55.90 $\pm$ 1.74}
& 50.54 $\pm$ 3.11
& 50.74 $\pm$ 2.60 \\

WuP$_o$
& \textbf{48.17 $\pm$ 1.99}
& 44.51 $\pm$ 1.66
& 45.22 $\pm$ 1.73
& 31.77 $\pm$ 5.93 \\

WuP$_f$
& 92.93 $\pm$ 3.18
& \textbf{94.64 $\pm$ 3.40}
& 96.63 $\pm$ 1.90
& 92.19 $\pm$ 6.90 \\

\midrule
\multicolumn{5}{c}{\textbf{Schema.org}} \\
\midrule

R@K
& 14.04 $\pm$ 2.69 & \textbf{23.28 $\pm$ 0.80} & 19.66 $\pm$ 1.06 & 11.06 $\pm$ 2.52 \\

SR@K
& 22.99 $\pm$ 2.24 & 31.70 $\pm$ 2.11  & \textbf{31.79 $\pm$ 0.96} & 19.23 $\pm$ 2.62 \\

MRR$_o$
& 7.86 $\pm$ 1.18 & \textbf{11.97 $\pm$ 0.46} & 11.02 $\pm$ 1.18 & 6.63 $\pm$ 1.67 \\

MRR$_f$
& 52.12 $\pm$ 4.86 & 49.72 $\pm$ 1.62 & 53.94 $\pm$ 3.56 & \textbf{55.39 $\pm$ 4.67} \\

WuP$_o$
& 21.75 $\pm$ 3.29 & \textbf{27.57 $\pm$ 1.28} & 24.92 $\pm$ 1.61 & 13.75 $\pm$ 4.52 \\

WuP$_f$
& 80.27 $\pm$ 1.10 & \textbf{92.32 $\pm$ 3.03} & 90.61 $\pm$ 2.52 & 85.71 $\pm$ 3.03 \\

\bottomrule
\end{tabular}
\vspace{-5mm}
\end{table}

Overall, all evaluated models demonstrate the ability to recover hidden taxonomy concepts. As expected, the larger and more diverse Schema.org taxonomy is substantially more challenging than the SemEval Environment taxonomy, leading to lower recovery and attachment scores across all models. Nevertheless, the relative ranking of the models remains largely consistent across both benchmarks. Mistral achieves the strongest overall performance across both datasets, obtaining the highest R@K, MRR$_o$, and WuP$_o$ scores on Schema.org while remaining highly competitive on SemEval. Llama3.2 performs best on the smaller SemEval taxonomy, achieving the highest R@K, S@K, and WuP$_o$, indicating strong concept recovery and structural placement. Qwen3 consistently produces competitive semantic recovery performance, achieving the highest S@K on Schema.org, suggesting a greater ability to generate semantically related concepts beyond exact lexical matches. DeepSeek-R1 generally obtains lower recovery scores and exhibits larger standard deviations across several metrics, indicating lower stability across different masking configurations.

\subsection{Human Evaluation}

%While the masked taxonomy expansion benchmark provides an objective assessment of concept recovery and structural attachment quality, it does not fully capture the semantic validity or usefulness of generated taxonomy expansions. In particular, 
LLMs may generate valid novel concepts that are absent from the ground truth. %and therefore cannot be rewarded by recovery-based metrics. 
To complement the automatic evaluation, we conduct a human assessment of the generated taxonomy expansions. %The evaluation is conducted on the SemEval Environment and Schema.org taxonomies. For each of the four generation models (Llama3.2, Mistral, Qwen3, and DeepSeek-R1), 
We evaluate the expanded taxonomy obtained from the run achieving the highest R@K. %among the six runs performed for each model. %Human evaluation is therefore performed on the same expanded taxonomies used in the automatic benchmark, ensuring that both evaluations assess identical generated outputs. 

For each taxonomy, we randomly sample 20\% of the eligible parent nodes, providing representative coverage while keeping the manual annotation effort feasible. Eligible parents correspond to non-root nodes that retain at least one visible child in the masked seed taxonomy. The same sampled parent nodes are evaluated across all four models. % to allow fair comparison under identical taxonomic contexts.

To avoid model-specific bias, the four models are anonymized. %as \emph{Model~1--4}. 
We further randomize the order of both the sampled parent nodes and the generated concepts. A web-based interface\footnote{\url{https://reltex-human-eval.vercel.app/}} shows the taxonomy context and allows consistent annotation.

For each sampled parent node, annotators are presented with its hierarchical path, its existing children, the accepted generated children, and, when applicable, the hidden gold children removed during masking (see Appendix~\ref{app:annotation_interface} for the snapshot of the annotation interface). The interface distinguishes between two evaluation settings. In \emph{recovery cases}, hidden gold children exist for the current parent, allowing both recovery and enrichment to be assessed. In \emph{novel enrichment cases}, no hidden gold child exists under the parent, and the evaluation focuses solely on the quality of the generated concepts. Each generated concept is independently evaluated according to the following criteria: \emph{Hierarchical Correctness (HC)}, \emph{Granularity Consistency (GC)}, \emph{Non-Redundancy (NR)}, \emph{Exact Recovery (ER)}, and \emph{Semantic Recovery (SR)}.

Three independent annotators evaluated all eight sample sets. Final labels are obtained using majority voting across the three annotations for each generated concept. The inter-annotator agreement was measured using Fleiss' $\kappa$, which is 0.695.

Table~\ref{tab:human_eval_majority} presents the final results. Expansion-quality metrics (HC, GC, and NR) are computed over all sampled parent nodes, whereas ER and SR are evaluated only on the subset of masked concepts. Despite being measured on this more restrictive evaluation set, the recovery scores demonstrate that the generated concepts frequently recover or closely match the hidden taxonomy concepts.

Overall, the generated expansions receive consistently high human judgments for hierarchical correctness, granularity consistency, and non-redundancy across both datasets, indicating that the accepted concepts are generally well positioned within the taxonomy and remain consistent with the abstraction level of the surrounding branch.

\begin{table}[t]
\centering
\scriptsize
\caption{Human evaluation results using majority voting across three annotators. }
\setlength{\tabcolsep}{4pt}
\renewcommand{\arraystretch}{1.08}

\begin{tabular}{@{}llccccc@{}}
\toprule
\textbf{Dataset} &
\textbf{Model} &
\multicolumn{3}{c}{\textbf{Expansion Quality}} &
\multicolumn{2}{c}{\textbf{Recovery}} \\
\cmidrule(lr){3-5}
\cmidrule(l){6-7}
&
&
\textbf{HC} &
\textbf{GC} &
\textbf{NR} &
\textbf{ER} &
\textbf{SR} \\
\midrule

\multirow{4}{*}{\shortstack[l]{SemEval\\Env}}
& Llama3.2
& 0.978
& 0.978
& \textbf{0.978}
& 0.200
& 0.300 \\

& DeepSeek-R1
& 0.960
& 0.960
& 0.920
& 0.200
& 0.300 \\

& Mistral
& 0.960
& 0.940
& 0.860
& \textbf{0.400}
& \textbf{0.400} \\

& Qwen3
& \textbf{0.980}
& \textbf{0.980}
& 0.940
& 0.300
& \textbf{0.400} \\
\midrule

\multirow{4}{*}{Schema.org}
& Llama3.2
& 0.898
& 0.932
& 0.864
& 0.103
& 0.144 \\

& DeepSeek-R1
& 0.988
& 0.988
& 0.940
& 0.076
& 0.143 \\

& Mistral
& 0.978
& \textbf{1.000}
& \textbf{1.000}
& 0.101
& \textbf{0.202} \\

& Qwen3
& \textbf{1.000}
& \textbf{1.000}
& 0.989
& \textbf{0.112}
& 0.178 \\
\bottomrule
\end{tabular}
\vspace{-5mm}
\label{tab:human_eval_majority}
\end{table}

On SemEval Environment, Qwen3 achieves the strongest performance in HC and GC, while Llama3.2 produces the highest non-redundancy score. Mistral obtains the best exact recovery and ties with Qwen3 for the highest SR, indicating a stronger ability to recover concepts removed during masking.

On Schema.org, Qwen3 achieves perfect HC and GC. Mistral reaches perfect GC and NR together with the highest SR, whereas DeepSeek-R1 also demonstrates strong expansion quality despite lower recovery performance.

\subsection{LLM-Based Taxonomy Evaluation}

We complement our evaluation using LITE~\cite{zhang2025lite}, an LLM-based framework for assessing the semantic and structural quality of taxonomies. LITE defines four measures: \textbf{(i)} \emph{Single Concept Accuracy (SCA):} evaluates the semantic clarity, validity, and coherence of individual taxonomy concepts; \textbf{(ii)} \emph{Hierarchy Relationship Rationality (HRR):} evaluates whether parent--child relations represent logically consistent hierarchical dependencies. \textbf{(iii)} \emph{Hierarchy Relationship Exclusivity (HRE):} evaluates whether sibling concepts are semantically distinct and non-overlapping. \textbf{(iv)} \emph{Hierarchy Relationship Independence (HRI):} evaluates the structural independence of concepts by measuring redundancy and semantic overlap among siblings.

Following LITE, we assess taxonomy quality using LLM-based judgments. While the original framework reports HRE and HRI separately, our implementation returns a single %hierarchy-level 
score jointly reflecting both criteria, which we report as the HRE/HRI score. LITE is further applied to the fully expanded taxonomies produced by the ReLTEx (instead of masking). %with recursive expansion enabled. 
As recursive expansion increases the number of generated nodes at each level, using the benchmark generation setting (\(k=5\) candidates per parent) results in large taxonomies with limited benefit. Therefore, we reduce the generation to \(k=3\). We evaluate both the seed taxonomy and the enriched taxonomy, allowing direct comparison before and after enrichment.

%The LITE framework employs GPT-based evaluators. 
We replace the proprietary evaluator in LITE with three open-source instruction-tuned LLMs deployed locally through Ollama\footnote{\url{https://ollama.com/}}: \texttt{Llama3.2}, \texttt{Qwen3:8B}, and \texttt{DeepSeek-R1:8B}. The enriched taxonomies are generated using \texttt{Mistral:7B}, selected as the generator due to its strongest overall performance in the masked taxonomy expansion benchmark. The three LLMs are used exclusively as evaluators. 
Tables~\ref{tab:lite_env} and~\ref{tab:lite_schema} report the LITE scores for the seed and the enriched taxonomy. Improvements over the seed taxonomy show that recursive enrichment preserves or enhances the quality while introducing new concepts.

% env taxo table
\begin{table}[h]

\centering
\small
\caption{LITE evaluation on the SemEval Environment.}
\label{tab:lite_env}
\setlength{\tabcolsep}{6pt}
\begin{tabular}{llccc}
\toprule
\textbf{Evaluator} & \textbf{Taxonomy} & \textbf{SCA} & \textbf{HRR} & \textbf{HRE/HRI} \\
\midrule
\multirow{2}{*}{Llama3.2}
    & Seed     &   \textbf{8.01}   &   8.19   &   7.74   \\
    & ReLTEx&  7.92    &   \textbf{8.24}   &  \textbf{7.75 }   \\
\midrule
\multirow{2}{*}{Qwen3}
    & Seed & \textbf{7.42}   &   6.82   &   7.83  \\
    & ReLTEx &  7.31    &  \textbf{6.96 }   & \textbf{8.88 }    \\
\midrule
\multirow{2}{*}{DeepSeek}
    & Seed     & 7.71 &  7.43 &  7.03 \\
    & ReLTEx &   \textbf{7.73} &  \textbf{8.78}  & \textbf{8.15} \\
\bottomrule
\end{tabular}
%\end{table}
\vspace{3mm}
% schema taxo table 
%\begin{table}[h]
\centering
\small
\caption{LITE evaluation on the Schema.org.}
\label{tab:lite_schema}
\setlength{\tabcolsep}{6pt}
\begin{tabular}{llccc}
\toprule
\textbf{Evaluator} & \textbf{Taxonomy} & \textbf{SCA} & \textbf{HRR} & \textbf{HRE/HRI} \\
\midrule
\multirow{2}{*}{Llama3.2}
    & Seed  & \textbf{8.18}   &    8.05  &   7.95 \\
    & ReLTEx & 7.94  &\textbf{8.29} &\textbf{8.30} \\
\midrule
\multirow{2}{*}{Qwen3}
    & Seed   &  8.35    &   \textbf{8.95}   &   8.59 \\
    & ReLTEx & \textbf{8.64}  & 8.68  & \textbf{8.76} \\
\midrule
\multirow{2}{*}{DeepSeek}
    & Seed     &   \textbf{8.19}   &   8.67   &   \textbf{8.05}   \\
    & ReLTEx &   8.07   &    \textbf{8.68}  &    7.94  \\
\bottomrule
\end{tabular}
\vspace{-4mm}
\end{table}

Across both taxonomies, the enriched taxonomies achieve high SCA, indicating that recursive expansion preserves semantic quality. HRR is generally maintained or improved after enrichment, suggesting that generated parent--child relations remain logically coherent. Similar trends are observed for HRE/HRI, with most evaluators assigning equal or higher scores to the enriched taxonomies, particularly on the Environment taxonomy.\\

The ablation study, qualitative error analysis, and expanded taxonomy statistics are provided in Appendix~\ref{appendix:ablation}, Appendix~\ref{error_analysis}, and Appendix~\ref{appendix:taxonomy_stats}, respectively.

\section{Conclusion} \label{sec:conclusion}

We presented ReLTEx, a framework for taxonomy enrichment using LLMs. Unlike existing taxonomy expansion approaches that assume candidate concepts to be already available, ReLTEx jointly performs concept generation and structure-aware validation through a classifier-based validation module and recursive expansion mechanism.
The proposed evaluation protocol combines automatic benchmarking, human assessment, and LLM-based taxonomy evaluation, providing complementary perspectives on taxonomy quality beyond concept recovery alone.

\section*{Limitations}

Our experiments rely on relatively compact open-source language models to enable efficient and reproducible experimentation. Larger and more capable models may further improve the quality of generated taxonomy expansions.

In addition, the proposed classifier is a learned approximation of hierarchical validity and may occasionally accept incorrect parent--child relations. Finally, recursive LLM-based taxonomy generation remains an emerging task without a standardized benchmark or directly comparable generation baseline, making comprehensive system-level comparisons difficult.
\section*{Ethical Considerations}

The human evaluation involved three volunteer annotators following predefined annotation guidelines. The collected annotations were used solely for research purposes, and no sensitive personal information is reported in this work.

The taxonomies used are publicly available benchmark resources and contain no personal or sensitive data.

Since ReLTEx relies on large language models for concept generation, it may inherit biases present in the underlying models and their training data, potentially affecting the generated concepts and their hierarchical placement. Although the proposed validation module mitigates these issues, some errors may still remain.

\bibliography{custom}

\appendix

\section{Input Prompts}
\subsection{Generation Prompt}
\label{app:generation-prompt}

The following prompt is used to generate candidate children for each expanded
taxonomy node. The placeholders enclosed in braces are dynamically replaced
during execution.

\begin{promptbox}
You are expanding a taxonomy.

Current node: {current_node}
Parent path: {parent_path}
Existing children of this node: {existing_children}

Task:
Suggest exactly {K} NEW suitable children of the current node.

Important:
- The new children MUST be at the SAME level of specificity as
  the existing children.
- Use the existing children as examples of the correct granularity.
- Each suggestion must be a child under "{current_node}".

Rules:
- Do NOT repeat any existing child.
- Do NOT suggest synonyms, rephrasings, or formatting variants.
- Do NOT suggest broader or more general concepts than the existing
  children.

Output format:
- Return exactly {K} names.
- One name per line.
- No numbering, explanations, or additional text.
- Do not return the parent path.
\end{promptbox}

\subsection{LLM-as-a-Judge Prompt}
\label{app:judge-prompt}

The following prompt is used by the LLM-based validator to assess each
generated candidate independently. The placeholders are populated using the
current taxonomy context.

\begin{promptbox}
You are validating a taxonomy expansion candidate.

Parent path: {parent_path}
Parent class: {parent}
Existing children of the parent: {existing_children}
Candidate child: {candidate}

Task:
Decide whether the candidate should be accepted as a reasonable
NEW child category under the parent in this taxonomy.

Guidelines:
1. Accept the candidate if it is primarily a meaningful type/category of the parent.
2. Reject the candidate if it is redundant with an existing child.
3. Reject the candidate if it is only loosely related to the parent.

Return ONLY valid JSON with exactly these fields:
{
  "accept": true or false,
  "is_type_of_parent": true or false,
  "is_redundant": true or false,
  "confidence": number between 0 and 1,
  "reason": "short explanation"
}
\end{promptbox}

A candidate is accepted only if the judge marks it as acceptable,
classifies it as a type of the parent, and does not identify it as
redundant:
\[
\textsc{Accept}(c)
=
a(c)
\land
t(c)
\land
\neg r(c),
\]
where \(a(c)\), \(t(c)\), and \(r(c)\) correspond to the
\texttt{accept}, \texttt{is\_type\_of\_parent}, and
\texttt{is\_redundant} fields, respectively.

\section{Human Evaluation Interface}
\label{app:annotation_interface}
\begin{figure*}[h]
    \centering
    \begin{subfigure}{0.48\textwidth}
        \centering
        \includegraphics[width=\linewidth]{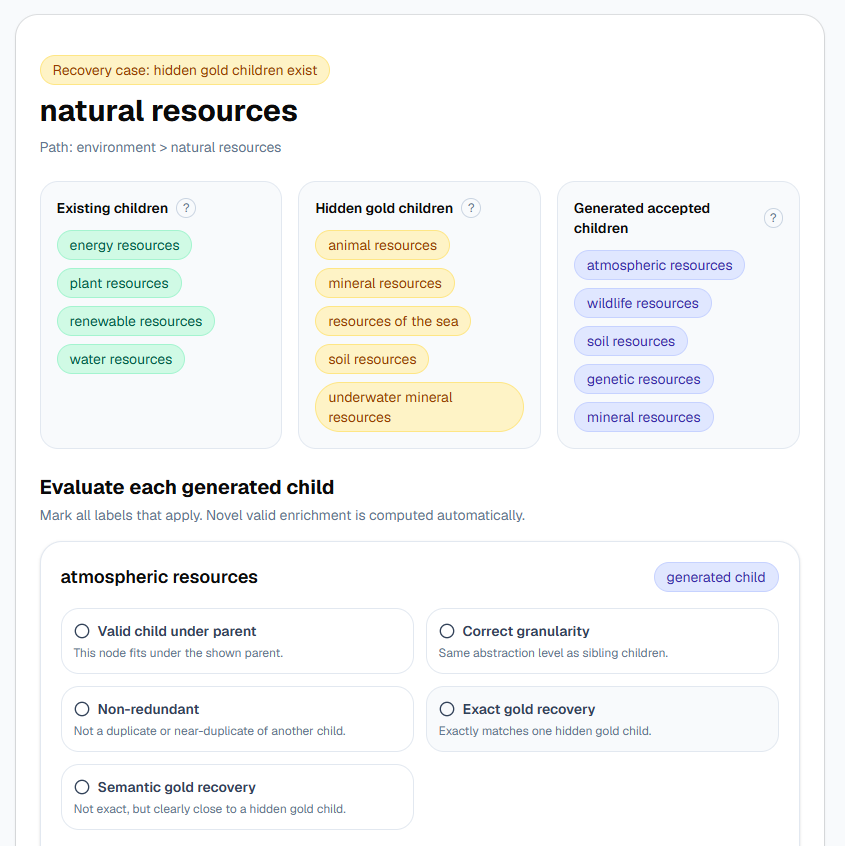}
        \caption{Recovery case (SemEval Environment).}
    \end{subfigure}
    \hfill
    \begin{subfigure}{0.50\textwidth}
        \centering
        \includegraphics[width=\linewidth]{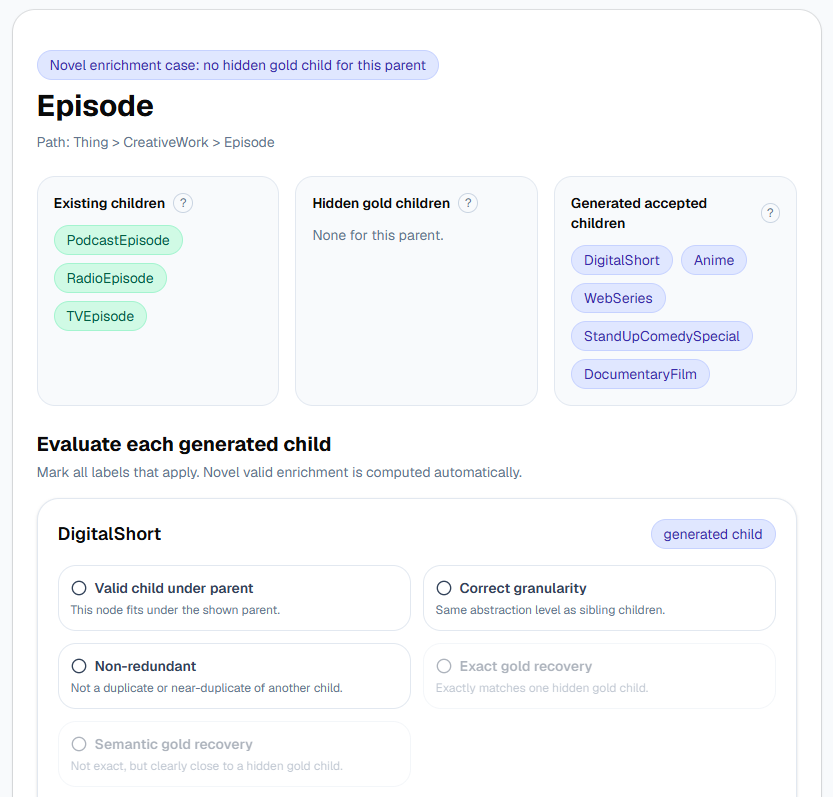}
        \caption{Novel enrichment case (Schema.org).}
    \end{subfigure}

    \caption{Snapshot of the annotation interface used in the human evaluation.}
    \label{fig:annotation-interface}
\end{figure*}

Figure~\ref{fig:annotation-interface} illustrates the annotation interface used throughout the human evaluation. The upper panel summarizes the local taxonomy context by displaying the hierarchical path, existing children, generated accepted children, and, when applicable, the hidden gold children. The lower panel presents the evaluation form for each generated concept. Recovery-specific criteria are automatically enabled only when hidden gold children are available, while novel enrichment cases evaluate only the validity and quality of newly generated concepts.

\section{Error Analysis} \label{error_analysis}

To better understand the remaining limitations of ReLTEx, we manually analyzed all parent--child relations that were accepted by our pipeline but rejected by the majority of human annotators. Rather than grouping errors according to the evaluation criteria, we categorized them based on their underlying semantic causes. Table~\ref{tab:error_analysis} presents representative examples from each category.

One category of remaining errors corresponds to \textbf{incorrect hierarchical placement}. In these cases, the generated concept is semantically valid but attached to an inappropriate parent within the taxonomy. For example, \textit{BankTransfer} was generated under \textit{PaymentCard}, although it represents an alternative payment method rather than a subtype of payment card. Similarly, \textit{overfishing} was proposed as a child of \textit{marine pollution}, despite representing a distinct environmental issue rather than a form of pollution.

Another category corresponds to \textbf{scope mismatch}. Here, the generated concept belongs to the correct semantic domain but is expressed at an incompatible level of abstraction. For instance, concepts such as \textit{Climate change mitigation strategies} and \textit{Climate change adaptation measures} were generated as children of \textit{Climate change}, mixing intervention strategies with environmental phenomena. Similar behavior was observed in Schema.org, where \textit{Gallery} was generated under \textit{MediaGallery}, although it is semantically more general than its parent.

We also observed cases of \textbf{semantic redundancy}, where the generated concept was effectively a near-synonym of an existing taxonomy concept. For example, \textit{TattooStudio} was generated despite the taxonomy already containing the semantically equivalent concept \textit{TattooParlor}, resulting in unnecessary duplication.

\begin{table*}[h]
\centering
\footnotesize
\caption{Representative semantic error categories identified during qualitative analysis.}
\label{tab:error_analysis}
\begin{tabular}{p{2.8cm}p{3.3cm}p{2.8cm}p{4.5cm}}
\toprule
\textbf{Generated Concept} &
\textbf{Existing Context} &
\textbf{Category} &
\textbf{Observation} \\
\midrule

\textit{BankTransfer} &
Parent: \textit{PaymentCard} &
Incorrect placement &
Not a subtype of the parent \\
\midrule

\textit{Gallery} &
Parent: \textit{MediaGallery} &
Scope mismatch &
Superclass of the parent \\
\midrule

\textit{TattooStudio} &
Sibling: \textit{TattooParlor} &
Semantic redundancy &
Near-synonymous existing concept \\
\midrule

\textit{Sho Emission Trading} &
-- &
Malformed concept &
Incomplete generated entity \\

\bottomrule
\end{tabular}
\end{table*}

Finally, a small number of errors correspond to \textbf{malformed concepts}, including incomplete or corrupted generations such as \textit{Sho Emission Trading}. These failures originate from generation artifacts rather than structural reasoning errors.

\section{Ablation Study}
\label{appendix:ablation}

To assess the impact of the main design choices in ReLTEx, we conduct ablation studies on three components of the framework: the prompting context, the validation strategy, and the validator acceptance thresholds. The experiments are performed on the SemEval Environment taxonomy using Mistral:7B as the generator model. As the goal is to isolate the contribution of each design choice rather than compare datasets, the ablation is conducted on this single representative benchmark.

Threshold calibration is performed through a sweep over a fixed set of generated candidates. Based on the resulting recovery performance, we select acceptance thresholds of $0.83$ for both semantic validators and $0.90$ for the classifier. These thresholds are used throughout all experiments reported in the paper.

We compare three prompting configurations:

\begin{itemize}
\item \textbf{Local}: provides only the path to the parent node together with its existing children.
\item \textbf{Parent Subtree}: provides the complete subtree rooted at the parent.
\item \textbf{Full Taxonomy}: provides the entire seed taxonomy as context.
\end{itemize}

For each prompting configuration, four validation strategies are evaluated:

\begin{itemize}
\item \textbf{Semantic V1}: validates candidates using cosine similarity between the parent and the generated child.
\item \textbf{Semantic V2}: extends Semantic V1 by additionally considering the taxonomy path and similarity to the parent's existing children.
\item \textbf{LLM Judge}: uses Llama3.2 to determine whether a generated candidate represents a valid, non-redundant child.
\item \textbf{Classifier}: the proposed DistilRoBERTa-based taxonomy validator.
\end{itemize}

Table~\ref{tab:ablation} summarizes the benchmark results.

Across all validation strategies, the \textit{Local} prompting configuration consistently achieves the highest R@K and SR@K. Restricting the context to the parent path and its existing children provides sufficient structural information while avoiding the additional noise introduced by larger contexts. Both the \textit{Parent Subtree} and \textit{Full Taxonomy} settings substantially reduce recovery performance, suggesting that exposing the language model to increasingly large portions of the taxonomy does not improve concept generation.

The comparison of validation strategies further supports the proposed classifier. Across all prompting contexts, it consistently achieves the highest R@K and SR@K. Both semantic validators produce slightly lower recovery scores by rejecting a small number of valid candidates, while Semantic V2 performs nearly identically to Semantic V1, indicating that the additional taxonomy context provides little benefit in this setting. The LLM Judge is the most conservative validator, substantially reducing recovery performance by rejecting many valid candidates together with incorrect ones.

\begin{table}[h]
\centering
\footnotesize
\caption{Results of the ablation study on the SemEval Environment taxonomy. Higher values indicate better performance.}
\label{tab:ablation}
\begin{tabular}{llcc}
\toprule
\textbf{Context} & \textbf{Validation Strategy} & \textbf{R@K} & \textbf{SR@K} \\
\midrule

\multirow{4}{*}{Local}
& Classifier & \textbf{34.62} & \textbf{42.31} \\
& Semantic V1 & 32.69 & 40.38 \\
& Semantic V2 & 32.69 & 40.38 \\
& LLM Judge & 15.38 & 19.23 \\

\midrule

\multirow{4}{*}{Parent Subtree}
& Classifier & \textbf{17.31} & \textbf{23.08} \\
& Semantic V1 & 15.38 & 21.15 \\
& Semantic V2 & 15.38 & 21.15 \\
& LLM Judge & 3.85 & 3.85 \\

\midrule

\multirow{4}{*}{Full Taxonomy}
& Classifier & \textbf{11.54} & \textbf{15.38} \\
& Semantic V1 & 9.62 & 13.46 \\
& Semantic V2 & 9.62 & 13.46 \\
& LLM Judge & 9.62 & 9.62 \\

\bottomrule
\end{tabular}
\end{table}

\section{Expanded Taxonomy Statistics}
\label{appendix:taxonomy_stats}

To provide a quantitative overview of the resulting enriched taxonomies, table~\ref{tab:taxonomy_statss} compares the original seed taxonomies with the enriched taxonomies produced by ReLTEx. To remain consistent with the LITE evaluation, we report the taxonomies generated using \textit{Mistral:7B}, which was selected as the generator after achieving the strongest overall performance in the masked taxonomy expansion benchmark.

\begin{table}[t]
\centering
\small
\caption{Statistics of the seed and enriched taxonomies produced by ReLTEx using \textit{Mistral:7B}. $|N|$ denotes the number of nodes, $|E|$ the number of edges, and $|D|$ the maximum depth.}
\label{tab:taxonomy_statss}
\setlength{\tabcolsep}{5pt}
\begin{tabular}{llccc}
\toprule
\textbf{Dataset} & \textbf{Taxonomy} & $|N|$ & $|E|$ & $|D|$ \\
\midrule

\multirow{2}{*}{SemEval Env}
& Seed    &  211 &  210  &  6  \\
& ReLTEx  &   4883  & 4882    &   7  \\

\midrule

\multirow{2}{*}{Schema.org}
& Seed    & 912  &  911   &   6  \\
& ReLTEx  &  38603   &  38602   &   7  \\

\bottomrule
\end{tabular}
\end{table}

The adaptive recursive expansion substantially increases the size of both taxonomies while introducing one additional hierarchy level. This indicates that ReLTEx primarily enriches existing branches rather than generating excessively deep hierarchies. The larger increase observed for Schema.org also reflects its broader semantic coverage, providing more opportunities for recursive concept generation than the smaller SemEval Environment taxonomy.

\label{sec:appendix}

\end{document}